\documentclass[letter]{spie}

\usepackage[utf8]{inputenc}
\usepackage{graphicx}
\usepackage{siunitx}
\usepackage{slashbox}
\usepackage{capt-of}
\usepackage{cleveref}
\usepackage{array}
\usepackage{calc}
\usepackage{amsmath}
\usepackage{subfig}
\usepackage{changepage}
\usepackage{multirow}
\title{Automatic Segmentation of Thoracic Aorta Segments in Low-Dose Chest CT}
\author{Julia M. H. Noothout\supit{a}, Bob D. de Vos\supit{a}, Jelmer M. Wolterink\supit{a}, Ivana I\v{s}gum\supit{a}
	\skiplinehalf
	\supit{a}Image Sciences Institute, University Medical Center Utrecht, Utrecht, The Netherlands\\
}

\begin{document}
	\maketitle 
	
	\begin{abstract}
	Morphological analysis and identification of pathologies in the aorta are important for cardiovascular diagnosis and risk assessment in patients. Manual annotation is time-consuming and cumbersome in CT scans acquired without contrast enhancement and with low radiation dose. Hence, we propose an automatic method to segment the ascending aorta, the aortic arch and the thoracic descending aorta in low-dose chest CT without contrast enhancement. Segmentation was performed using a dilated convolutional neural network (CNN), with a receptive field of $131\times131$ voxels, that classified voxels in axial, coronal and sagittal image slices. To obtain a final segmentation, the obtained probabilities of the three planes were averaged per class, and voxels were subsequently assigned to the class with the highest class probability. Two-fold cross-validation experiments were performed where ten scans were used to train the network and another ten to evaluate the performance. Dice coefficients of $0.83 \pm 0.07$, $0.86 \pm 0.06$ and $0.88 \pm 0.05$, and Average Symmetrical Surface Distances (ASSDs) of $2.44 \pm 1.28$, $1.56 \pm 0.68$ and $1.87 \pm 1.30$ mm were obtained for the ascending aorta, the aortic arch and the descending aorta, respectively. The results indicate that the proposed method could be used in large-scale studies analyzing the anatomical location of pathology and morphology of the thoracic aorta.
	\end{abstract}

	
	\section{Introduction}
	Accurate segmentation of the aorta in CT can be used to analyze morphology and detect pathology such as atherosclerotic plaque and aneurysms \cite{erbel2006aortic}\spacefactor\sfcode`\.{}. Moreover, the location of specific shape changes or pathology in the aorta is relevant for diagnosis and risk assessment in patients\cite{erbel2006aortic,french1996atherosclerotic}\spacefactor\sfcode`\.{}. However, manual annotation of the aorta and its subdivision into segments is time-consuming and cumbersome, especially in low-dose chest CT scans where a lack of contrast enhancement leads to low soft-tissue contrast and acquisition with a low radiation dose may result in high levels of image noise.
	
	Thus far, several methods have been developed for automatic segmentation of the aorta in low-dose non-contrast-enhanced chest CT scans. Kurugol et al. \cite{kurugol2012aorta} used Hough transforms on computed oblique and on axial slices to segment the aorta. Using the results of the Hough transforms, the surface of the aorta was reconstructed and thereafter, the segmentations were refined using level sets. Xie et al.\cite{xie2014automated} proposed an algorithm that iteratively fits cylinders of varying lengths to track the aorta in the image. The cylindrical model is fit in the image space defined by previously segmented organs surrounding the aorta, such as the lungs and trachea. Finally, segmentation of the aorta is refined using local image intensities. I\v{s}gum et al. \cite{isgum2009multi} employed atlas-based registration that locally combines atlases based on the registration success of each atlas.
	
	Even though these methods generally obtain good results, they require either segmentation of neighboring organs\cite{kurugol2012aorta, xie2014automated} (e.g. lungs or airways) or manual tuning of parameters, such as the atlases used \cite{isgum2009multi}\spacefactor\sfcode`\.{}. Furthermore, existing automatic methods only segment the complete thoracic aorta and do not subdivide it into segments. Therefore, we propose an automatic method to segment the ascending aorta, aortic arch and descending aorta in low-dose, non-contrast-enhanced chest CT. We employ a dilated convolutional neural network (CNN) that analyzes axial, coronal and sagittal CT slices to classify voxels into one of the three aortic segments or background. The results obtained in each image plane are merged to provide the final segmentation result. Unlike previous methods that exploit the expected shape of the aorta, CNNs are capable of using CT images as input and automatically acquire hierarchical feature representations needed for the segmentation task.
	
	\section{Data}
	This study included 24 low-dose chest CT scans, randomly chosen from a set of baseline scans acquired in the National Lung Screening Trial (NLST)\cite{national2011national}\spacefactor\sfcode`\.{}. All scans were acquired during inspiratory breath-hold in supine position with the arms elevated above the head and included the outer rib margin at the widest patient dimension. The selected scans were acquired on seven different scanners of three major CT scanner vendors (GE, Siemens and Philips). Depending on patient weight, a tube voltage of 120 kVp or 140 kVp and a tube current ranging between 30 and 160 mAs were used. Scans were made using an axial reconstruction with an in-plane resolution varying between 0.46 and 0.86 mm, a slice thickness varying between 1.25 and 4.00 mm, and a slice spacing varying between 0.63 and 3.00 mm. No contrast enhancement or ECG-triggering was applied.
	
\begin{figure}	
		\centerline{
			\subfloat{\includegraphics[width=4.5cm, height=3.7cm, trim={0cm 0cm 0cm 1.7cm}, clip]{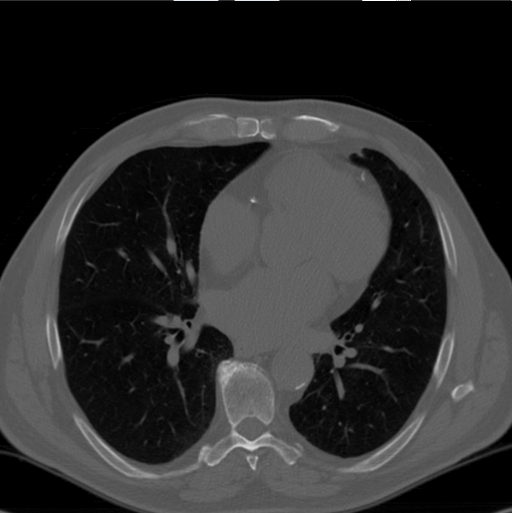}}
			\hspace{0.3em}
			\subfloat{\includegraphics[width=4.5cm, height=3.7cm, trim={0cm 0cm 0cm 2.5cm}, clip]{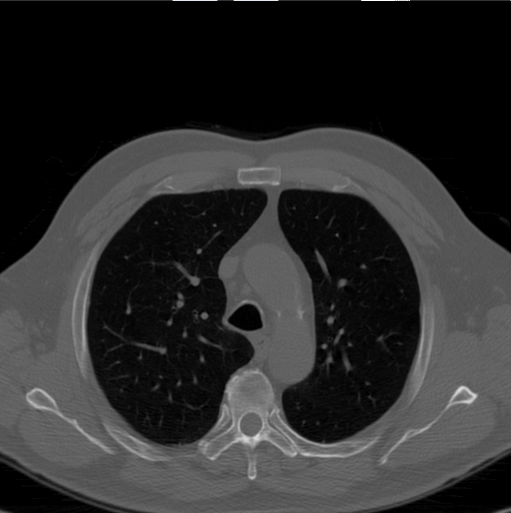}}
			\hspace{0.3em}
			\subfloat{\includegraphics[width=4.5cm, height=3.7cm, trim={3.7cm 0cm 4.1cm 2cm}, clip]{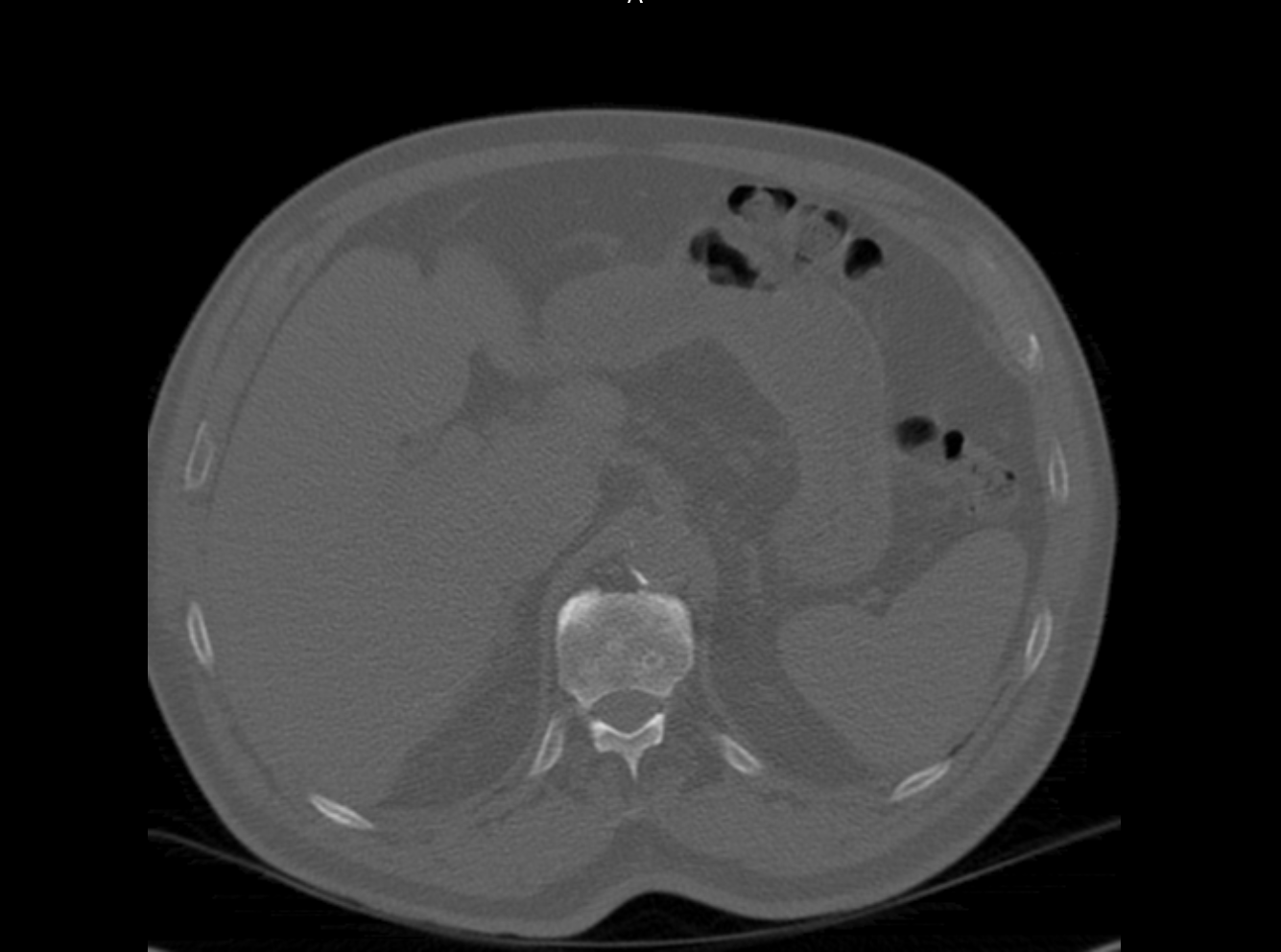}}}
		\vspace{0.3em}
		
		\centerline{
			\subfloat{\includegraphics[width=4.5cm, height=3.7cm, trim={1.8cm 0cm 1.8cm 1.7cm}, clip]{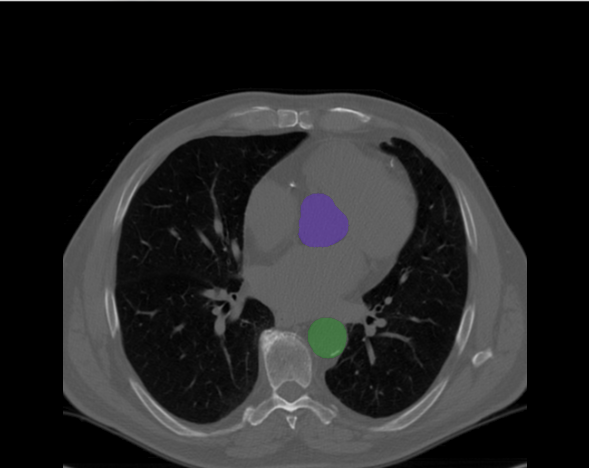}}
			\hspace{0.3em}
			\subfloat{\includegraphics[width=4.5cm, height=3.7cm, trim={1.7cm 0cm 1.8cm 2.3cm}, clip]{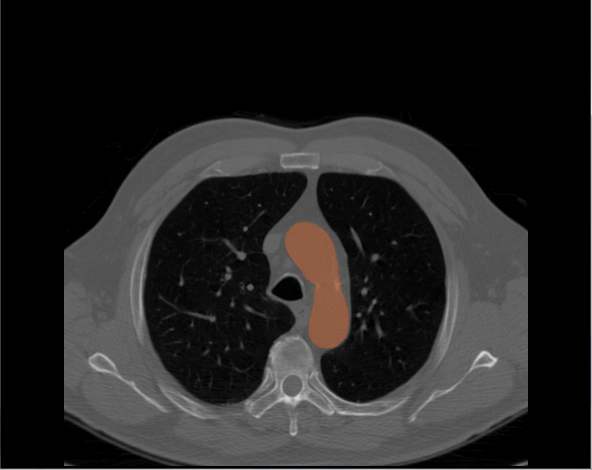}}
			\hspace{0.3em}
			\subfloat{\includegraphics[width=4.5cm, height=3.7cm, trim={1.6cm 0cm 1.5cm 1.3cm}, clip]{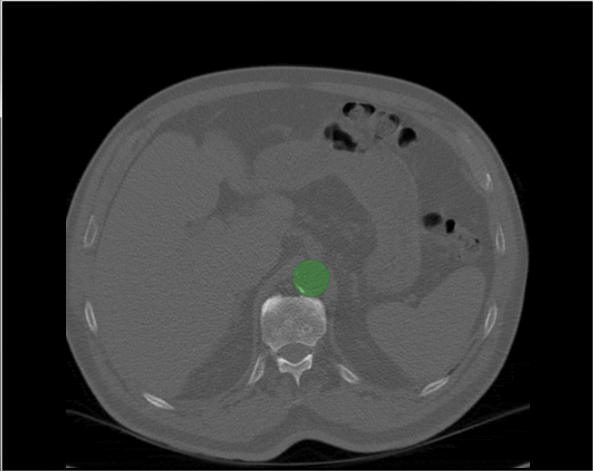}}}
		\vspace{0.3em}
		
		\caption{CT slices (top) and the corresponding manual annotation (bottom) showing the ascending aorta (purple), aortic arch (orange) and descending aorta (green). The images illustrate the difficulty of the task due to a lack of intravenous contrast and the presence of image noise.}
		\label{man_segm}	
	\end{figure}

	Reference annotations were obtained by manual voxel painting of the aorta in the axial plane (Fig. \ref{man_segm}). Specific labels were assigned to the ascending aorta, the aortic arch and the descending aorta. The aortic arch was defined as the section of the aorta where the ascending and descending aorta are connected\cite{de2017convnet}\spacefactor\sfcode`\.{}. The ascending aorta was defined from the aortic root up to the aortic arch and the descending aorta was defined from the aortic arch down to the last axial slice of a scan.

	\section{Method}
\begin{figure}[t]
		\centering	
		\includegraphics[scale = 0.72]{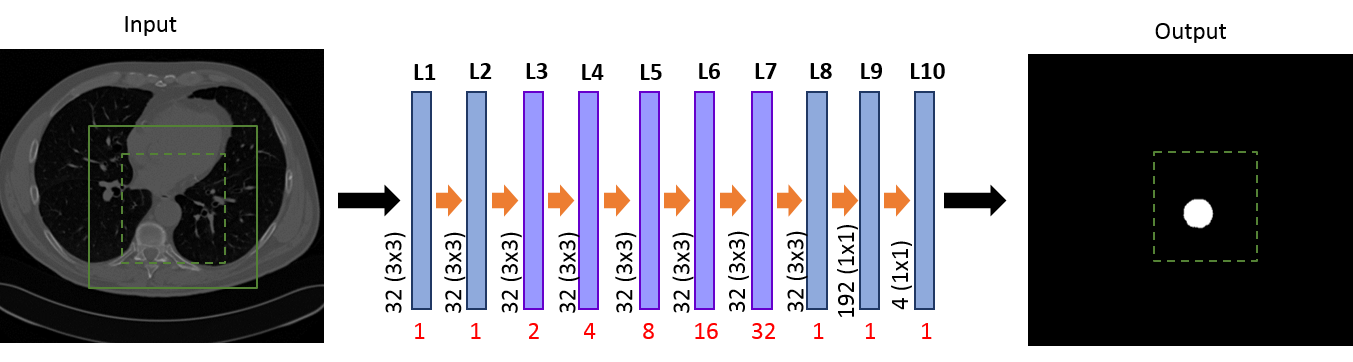}
		\caption{Architecture of the dilated CNN, containing ten convolutional layers with dilation factors (indicated in red) increasing from 1 in the first layer to 32 in the seventh layer. Each convolutional layer has 32 $3\times3$ filters, except layers 9 and 10, which are fully connected layers implemented as $1\times1$ convolutions. Every convolutional layer is followed by a rectified linear unit (ReLU) activation function, except for the output layer, which is followed by a softmax function. The output layer contains four output units, one for every class: the ascending aorta, the aortic arch, the descending aorta and the background. During training, sub-images of $281\times281$ voxels (green square) are used as input of which $151\times151$ voxels (dashed green square) are classified.}
		\label{networkfig}	
	\end{figure}

 	To segment the aorta, a CNN is trained to assign a class label to every voxel in a scan based on classification in three orthogonal image slices. A lack of contrast enhancement in scans leads to homogeneous image intensities, especially around the ascending aorta (Fig. \ref{man_segm}). Hence, the precise location of the aorta has to be inferred from a larger image context. To use a large receptive field and to keep the number of parameters low, a dilated CNN (Fig. \ref{networkfig}) is employed. It has a similar architecture as the networks described by Wolterink et al.\cite{wolterink2016dilated} and Yu et al.\cite{yu2015multi} and analyzes 2D image slices using ten convolutional layers. The size of the receptive field is set to $131\times131$ voxels but, due to increasing dilation factors in subsequent convolutional layers, the network only contains 72,643 trainable parameters. Dropout\cite{srivastava2014dropout} (p=0.5) and batch normalization\cite{ioffe2015batch} are applied to the fully connected layers to prevent overfitting. To compensate for varying in-plane resolutions, prior to analysis all scans are resized to an isotropic resolution of 1 mm.

	The CNN is purely convolutional, thus it is able to analyze images of a variable size. Therefore, during training, batches containing sub-images in the axial, sagittal and coronal planes are analyzed, and during testing, full slices padded with 65 voxels in all directions are used as input. Moreover, all slices from the axial, sagittal and coronal planes of a scan are analyzed. This results in three multi-class 3D probability maps: one map for each plane orientation. A final probability map is determined by averaging these three multi-class probability maps. Results are resampled from isotropic resolution to the original image resolution using trilinear interpolation and subsequently, each voxel is assigned the class with the highest class probability. To prevent small isolated clusters of voxels being segmented, only the largest component for each class is included in the final segmentation.
	
	Performance of the trained network was evaluated by the Dice coefficient as an overlap measure between automatically obtained and reference segmentations. Furthermore, the Average Symmetrical Surface Distance (ASSD) was computed to evaluate the segmentation along the aortic boundary. The evaluation was performed for each class separately.
	
	\section{Experiments and Results}
	Two-fold cross-validation experiments were performed with 24 CT scans. In each experiment, ten scans were used for training and another ten scans were used for testing the method. The remaining four scans were used as validation set to ensure no overfitting occurred during training. Unlike in the experiments presented by Wolterink et al. \cite{wolterink2016dilated}\spacefactor\sfcode`\.{} where categorical cross-entropy was used as a loss function, the current work employed the Dice coefficient as a loss function to address class imbalance in our data set\cite{milletari2016v}\spacefactor\sfcode`\.{}. The Adam optimization algorithm\cite{kingma2014adam} (learning rate = 0.001) was used to optimize the network parameters during 250,000 training iterations. In each iteration, a mini-batch containing 16 randomly sampled $281\times281$ sub-images from the three planes was provided to the network. The same hyperparameters were used for both cross-validation experiments.

	%\captionsetup[table]{aboveskip=-5pt}  
	%\setlength{\textfloatsep}{-2pt}
	\begin{table}[h]
		\caption{Segmentation task, number of training and test images, and the Dice coefficients and ASSDs obtained with the proposed segmentation method and as reported in previous work. To compare our results with previous work, the multi-class network was evaluated by merging the three aorta classes (ascending aorta, aortic arch and descending aorta) to one aorta class (thoracic aorta). In addition, the multi-class network was retrained using two classes: aorta and background.}
		\label{tbl_diceASSD}
		\begin{center}		
			\begin{tabular}{l l|c c c c}
				Method & Segmentation task & Training images & Test images &  Dice & ASSD (mm) \\ \hline
				Multi-class & Ascending aorta & 10 & 10 & $0.83 \pm 0.07$ & $2.44 \pm 1.28$ \\
						    & Aortic arch &  10 & 10 & $0.86 \pm 0.06$ & $1.56 \pm 0.68$  \\
							& Descending aorta & 10 & 10 & $0.88 \pm 0.05$ & $1.87 \pm 1.30$  \\
		   			 	    & Thoracic aorta & 10 & 10 & $0.89 \pm 0.05$ & $1.67 \pm 1.02$ \\
				Two-class & Thoracic aorta & 10 & 10 & $0.91 \pm 0.04$ & $1.32 \pm 0.85$  \\
				\hline
				Kurugol et al. \cite{kurugol2012aorta} & Thoracic aorta & - & 45 & $0.93 \pm 0.01$ & - \\
				Xie et al. \cite{xie2014automated} & Thoracic aorta & 20 & 60 & $0.93 \pm 0.01$ & $1.39 \pm 0.19$ \\
				Isgum et al. \cite{isgum2009multi} & Thoracic aorta & 15 & 14 & $0.87 \pm 0.03$ & - 					
			\end{tabular}
		\end{center}	
	\end{table}  

	Table \ref{tbl_diceASSD} lists the average ($\pm$ standard deviation) Dice coefficients and ASSDs achieved on the test scans. The best performance was obtained for the descending aorta, both in terms of Dice coefficient and ASSD. In contrast, the lowest performance was obtained for the ascending aorta.
	
	Previously described methods only segmented the aorta as a whole. To compare the performance of the proposed method with previous work, we retrained the network to perform two-class classification (aorta and background). This two-class segmentation network obtained slightly better results than the network trained for multi-class segmentation of the aorta (Table \ref{tbl_diceASSD}). Compared with other methods, both the multi-class and the two-class segmentation networks obtained competitive results. However, due to differences in used data and evaluation procedures among studies results can not be directly compared, but should be used as indication of the performance.
	
\begin{figure}[t]	
		\centerline{
			\subfloat[Reference segmentation]{\includegraphics[width=4.5cm, height=4.5cm, trim={0cm 0.6cm 0cm 0.4cm}, clip]{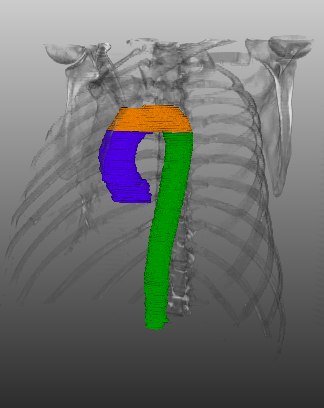}}
			\hspace{0.3em}
			\subfloat[Multi-class CNN segmentation]{\includegraphics[width=4.5cm, height=4.5cm, trim={0cm 0cm 0.2cm 0.5cm}, clip]{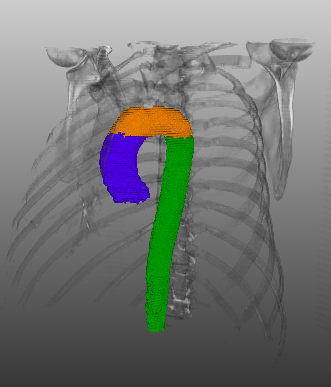}}
			\hspace{0.3em}
			\subfloat[Two-class CNN segmentation]{\includegraphics[width=4.5cm, height=4.5cm, trim={0.5cm 0.5cm 0cm 0.7cm}, clip]{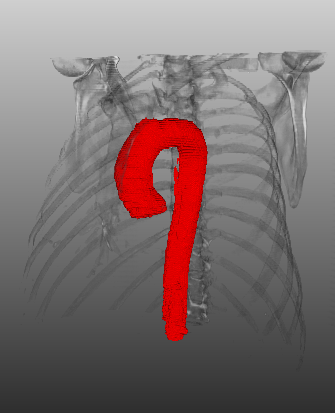}}}
		\vspace{0.2em}
		
		\centerline{
			\subfloat{\includegraphics[width=4.5cm, height=4.5cm, trim={0cm 0.6cm 0.7cm 0.5cm}, clip]{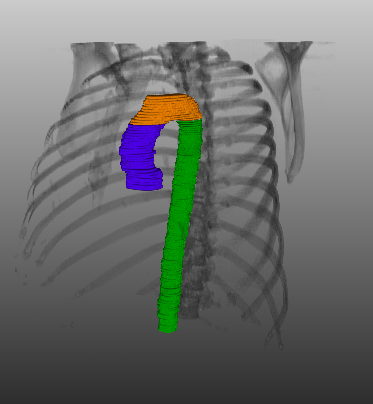}}
			\hspace{0.3em}
			\subfloat{\includegraphics[width=4.5cm, height=4.5cm, trim={0cm 1.3cm 0.6cm 0.4cm}, clip]{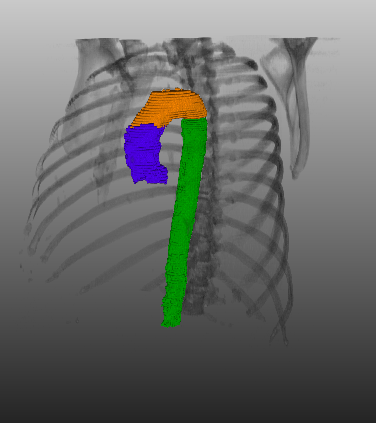}}
			\hspace{0.3em}
			\subfloat{\includegraphics[width=4.5cm, height=4.5cm, trim={0cm 1cm 0.9cm 0.3cm}, clip]{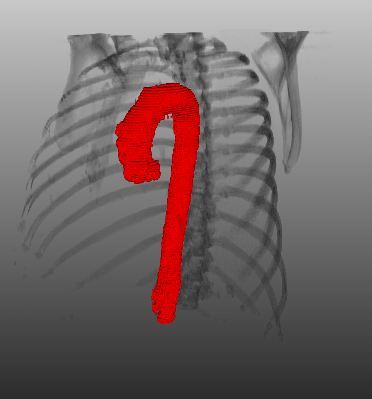}}}
		\vspace{0.2em}
	
		\centerline{
			\subfloat{\includegraphics[width=4.5cm, height=4.5cm, trim={0.1cm 0.9cm 0.4cm 0.6cm}, clip]{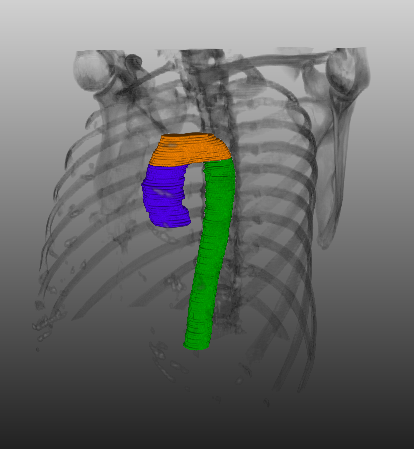}}
			\hspace{0.3em} 	
			\subfloat{\includegraphics[width=4.5cm, height=4.5cm, trim={0cm 1cm 0.2cm 0.5cm}, clip]{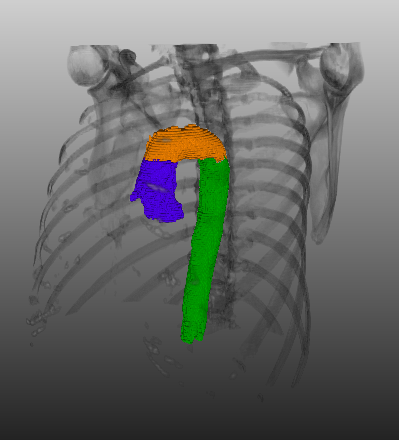}}
			\hspace{0.3em}
			\subfloat{\includegraphics[width=4.5cm, height=4.5cm, trim={0.3cm 1.0cm 0.4cm 0.8cm}, clip]{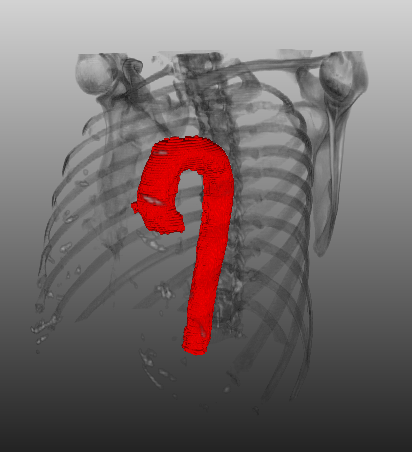}}}
		
		\caption{Segmentations of three test images (rows) obtained with the proposed classification method. First column: The reference segmentations with the ascending aorta (purple), the aortic arch (orange) and the descending aorta (green). Second column: The automatic segmentations obtained with the multi-class CNN. Third column: The automatic segmentations obtained with the two-class CNN.}
		\label{fig_results}	
	\end{figure}

	\newpage
	Fig. \ref{fig_results} shows segmentations obtained with the presented network trained for the multi-class and two-class segmentation problem. Results show that inaccuracies in classification may occur on the interface between different aortic segments. Nevertheless, no large differences are seen between automatic segmentations obtained with the multi-class and the two-class segmentation networks.  

	\captionsetup[subfigure]{labelformat=empty, position=top}

	\section{Discussion and Conclusion}
	We have presented a method for automatic segmentation of the ascending aorta, the aortic arch and the thoracic descending aorta in low-dose, non-contrast-enhanced chest CT scans using a purely convolutional neural network with dilated convolutions. The network is able to accurately segment the aorta. In addition, the proposed method obtained similar results as a network classifying voxels in only two classes (aorta and background). Moreover, the obtained results are on par with the results obtained in previous studies that only segment the aorta as a whole \cite{kurugol2012aorta, xie2014automated, isgum2009multi}\spacefactor\sfcode`\.{}.
	
	Dilated convolutions enable analysis with a large receptive field while keeping the number of network parameters low. This large receptive field allowed accurate detection of the aorta based on context information. Furthermore, because the network is purely convolutional, it is able to analyze images of a variable size. Hence, full slices could be segmented during testing even though the network was only trained with sub-images. On average, the segmentation took only 61.5 seconds per scan, making it suitable for application in studies including large numbers of images.
	
	In this study, results were least accurate in the ascending aorta. This is similar to other studies that achieved the least accurate segmentation results near the aortic root\cite{xie2014automated}\spacefactor\sfcode`\.{}. In low-dose non-contrast-enhanced chest CT it is often very difficult to outline the aortic root due to low soft-tissue contrast. A previous study reported substantial inter-observer disagreement in that region \cite{isgum2009multi}\spacefactor\sfcode`\.{}. Visual inspection of the here obtained automatic results revealed occasional inaccuracies just near the aortic root. Furthermore, in this study, results showed that segmentation of the descending aorta below the lungs was sometimes difficult due to high levels of image noise. Nevertheless, the employed CNN was able to overall segment the aorta accurately.
	
	Our experiments showed that the overall aorta segmentation was slightly more accurate when using two-class segmentation than when using multi-class segmentation. This could be due to differences between the two tasks. First, the multi-class segmentation problem can be considered more complex than the two-class segmentation problem, and may require more labeled training samples. Second, the emphasis on accurate overall aorta segmentation is stronger in the two-class task than in the multi-class task due to the Dice loss function used. In future work, the loss function for the multi-class segmentation task could potentially be adapted to correct for this. 
	
	In this study, three image planes were analyzed independently. In our future work we will investigate whether a different way of merging the results from the three image planes or alternatively extending analysis to 3D might be beneficial. In addition, to ensure that the method is able to accurately segment the aorta in images showing large anatomical variability (e.g. atherosclerotic plaque, aneurysm) and in images acquired with a large range of image acquisition parameters (different hospitals, scanners and reconstruction parameters), we will increase the size of the dataset to ensure the presence of a large range of variability in the training and test images. Given that clinical analysis of the morphology of the aorta is routinely performed on contrast-enhanced images, we will extend the evaluation to clinically acquired contrast enhanced chest CT scans.
	
	\section{new or breakthrough work to be presented}
	A method for automatic segmentation of the thoracic aorta into the ascending aorta, the aortic arch and the descending aorta in low-dose, non-contrast-enhanced chest CT scans is presented. This could be a first step towards large-scale studies analyzing anatomical location of pathology and morphology of the thoracic aorta.
	
	\acknowledgments 
	The authors thank the National Cancer Institute for access to NCI's data collected by the National Lung Screening Trial. The statements contained herein are solely those of the authors and do not represent or imply concurrence or endorsement by NCI.
	\bibliographystyle{spiebib}
	\bibliography{bibliography}

\end{document}